\documentclass{bmvc2k}
\usepackage{wrapfig,booktabs}

\title{Adversarial Robustness Comparison of \\Vision Transformer and MLP-Mixer to CNNs}

\addauthor{Philipp Benz$^*$}{phibenz@gmail.com}{1}
\addauthor{Soomin Ham$^*$}{smham@kaist.ac.kr}{1}
\addauthor{Chaoning Zhang$^*$}{chaoningzhang1990@gmail.com}{1}
\addauthor{Adil Karjauv}{mikolez@gmail.com}{1}
\addauthor{In So Kweon}{iskweon77@kaist.ac.kr}{1}

\addinstitution{
 Robotics and Computer Vision Lab\\
 Korea Advanced Institute of Science and Technology (KAIST)\\
 Daejeon, South Korea \\ \vspace{3mm}
 \small $^*$Equal contribution
}

\runninghead{Benz \etal}{Adv. Robustness Comparison of ViT and MLP-Mixer to CNNs}

\def\etal{\emph{et al}\bmvaOneDot}
\def\ie{\emph{i.e}\bmvaOneDot}
\def\vs{\emph{vs}\bmvaOneDot}

\begin{document}

\maketitle

\begin{abstract}
Convolutional Neural Networks (CNNs) have become the {\em de facto gold standard} in computer vision applications in the past years. Recently, however, new model architectures have been proposed challenging the {\em status quo}. The Vision Transformer (ViT) relies solely on attention modules, while the MLP-Mixer architecture substitutes the self-attention modules with Multi-Layer Perceptrons (MLPs). Despite their great success, CNNs have been widely known to be vulnerable to adversarial attacks, causing serious concerns for security-sensitive applications. Thus, it is critical for the community to know whether the newly proposed ViT and MLP-Mixer are also vulnerable to adversarial attacks. To this end, we empirically evaluate their adversarial robustness under several adversarial attack setups and benchmark them against the widely used CNNs. Overall, we find that the two architectures, especially ViT, are more robust than their CNN models. Using a toy example, we also provide empirical evidence that the lower adversarial robustness of CNNs can be partially attributed to their shift-invariant property. Our frequency analysis suggests that the most robust ViT architectures tend to rely more on low-frequency features compared with CNNs. Additionally, we have an intriguing finding that MLP-Mixer is extremely vulnerable to universal adversarial perturbations. Code: \url{https://github.com/phibenz/robustness_comparison_vit_mlp-mixer_cnn}. 
\end{abstract}

\section{Introduction}
\label{sec:intro}
Convolutional Neural Networks (CNNs)~\cite{lecun2015deep} have been the {\em gold standard} architecture in computer vision. In Natural Language Processing (NLP), however, attention-based transformers are the dominant go-to model architecture~\cite{devlin2018bert,radford2018improving,radford2019language}. Various attempts have been made to apply such transformer architectures to computer vision tasks~\cite{child2019generating,parmar2018image,ramachandran2019stand,chen2020generative}. A breakthrough moment was achieved with the advent of the Vision Transformer (ViT)~\cite{dosovitskiy2021an}, presenting a transformer architecture achieving comparable performance to state-of-the-art CNN architectures. Recently, another MLP-Mixer model architecture~\cite{tolstikhin2021mlpmixer} which does not rely on convolutions or self-attention, has been presented competing with CNN and ViT. Following~\cite{tolstikhin2021mlpmixer}, MLP-Mixer is termed Mixer in the remainder of this work for simplicity. 

Despite the success of CNNs, they are widely known to be vulnerable to adversarial examples~\cite{szegedy2013intriguing,goodfellow2014explaining} whose small additive perturbations of the input cause the CNN to misclassify a sample. This vulnerability causes serious concerns in security-sensitive applications, and thus it is also important to know whether the recently proposed ViT and Mixer are also vulnerable to adversarial attacks. 
This work sets out to evaluate the adversarial vulnerability of ViT and Mixer architectures and compare their robustness against the CNN models. Therefore, a wide range of adversarial attack methods has been adopted for a comprehensive study. 
Specifically, first, the performance of the different architectures is compared under the white-box attack, where an adversary has full knowledge of the model parameters to attack. Overall, the two newly proposed architectures, especially ViT, exhibit significantly higher robustness than CNNs against adversarial examples. We further compare their robustness under both query-based and transfer-based black-box attacks. In both cases, we observe a similar trend that among the three explored architectures, ViT is the most robust architecture while CNN is the least robust. 

To facilitate the understanding of why CNN is more vulnerable, we design a toy task of binary classification where each class is only represented by a single image. The image from each class has either a vertical or horizontal black stripe in the center. We find that the adversarial example for a CNN exhibits repetitive stripes over the image, while that of an FC network mainly exhibits a single stripe in the center. This observation indicates that the vulnerability of a CNN might be partially attributed to the fact that a CNN, which exploits local connections and shared weights by convolving kernels, is shift-invariant~\cite{zhang2019making,lenc2015understanding}. We also attempt to provide an analysis from the perspective of frequency, investigating whether the different model architectures are biased toward learning more high-frequency or low-frequency features. We find that the ViT seems to learn more low-frequency features, while the CNN is biased towards high-frequency features. Finally, we also investigate their robustness against common corruptions~\cite{hendrycks2019benchmarking} and universal adversarial perturbations~\cite{moosavi2017universal}.

\section{Related Work}
{\bf Beyond CNNs for vision applications.} In Natural Language Processing (NLP), transformers~\cite{vaswani2017attention}, which are solely based on the attention mechanisms, are the predominant model architecture~\cite{devlin2018bert,radford2018improving,radford2019language}. In contrast, CNNs have been the {\em de facto} standard in deep learning for vision applications, while the application of transformers to vision tasks is an emerging trend~\cite{child2019generating,parmar2018image,ramachandran2019stand,chen2020generative}. The Vision Transformer (ViT)~\cite{dosovitskiy2021an} was recently introduced, demonstrating that transformers can achieve state-of-the-art performance, by sequencing the images into patches and pre-training the model on large amounts of data. To address the data issue, DeiT~\cite{touvron2020training} introduced a teacher-student strategy specific to transformers and trained a transformer architecture only on the ImageNet-1K dataset. Concurrently, the T2T-ViT had been proposed~\cite{yuan2021tokens} introducing an advanced Tokens-to-Tokens strategy. Further works are attempting to extend the ViT architecture to increase the efficiency and performance of transformer architectures~\cite{chu2021conditional,han2021transformer,liu2021swin,wu2021cvt}. 
ViTs have further been explored beyond the task of image classification~\cite{wang2021pyramid,chen2021transunet,jiang2021transgan,neimark2021video,he2021transreid}.
Tolstikhin~\etal~\cite{tolstikhin2021mlpmixer} challenge the {\em status-quo} of convolutions and attention in current computer vision models and proposes MLP-Mixer, a pure Multi-Layer Perceptron (MLP)-based architecture that separates the per-location operations and cross-location.

{\bf Adversarial attacks and robustness.} CNNs are commonly known to be vulnerable to adversarial examples~\cite{szegedy2013intriguing,goodfellow2014explaining,kurakin2016adversarial}, which has prompted numerous studies on model robustness under various types of adversarial attacks. Depending on the accessibility to the target model, adversarial attacks can be divided into white box ones~\cite{goodfellow2014explaining,moosavi2016deepfool,carlini2017towards,madry2017towards} that require full access to the target model, query-based black-box attacks~\citep{chen2017zoo,papernot2017practical,ilyas2018prior,ilyas2018black,guo2019simple,tu2019autozoom,shi2019curls,rahmati2020geoda}, and transfer-based black-box attacks~\cite{dong2018boosting,xie2019improving,liu2016delving,tramer2017ensemble,wu2020skip,guo2020backpropagating,inkawhich2020perturbing}. Adversarial attacks can be divided into image-dependent ones~\cite{goodfellow2014explaining,moosavi2016deepfool,carlini2017towards,madry2017towards,rony2019decoupling} and universal ones~\cite{moosavi2017universal,mopuri2018nag,zhang2019cd-uap,zhang2020understanding,benz2020double,zhang2021survey}. Specifically, contrary to image-dependent attacks, a single perturbation, \ie universal adversarial perturbation (UAP) exists to fool the model for most images~\cite{zhang2021survey}. Based on the above various attack methods, this work empirically investigates and compares the adversarial robustness of ViT and Mixer architectures to CNN models. The vulnerability of transformers in the context of NLP tasks has also been investigated~\cite{hsieh2019robustness,jin2020bert,shi2020robustness,hendrycks2020pretrained,li2020bert,garg2020bae,hao2020self}. However, our work mainly focuses on the empirical robustness evaluation of the three architectures, namely CNN, ViT, and Mixer, in image classification.

\textbf{Concurrent works on similar topics.} Recently, there are a line of works~\cite{shao2021adversarial,bhojanapalli2021understanding,mahmood2021robustness,paul2021vision,mao2021rethinking,naseer2021intriguing,aldahdooh2021reveal,naseer2021improving,guo2021gradient,wei2021towards} that have investigated transformers from the perspective of adversarial robustness. Specifically,~\cite{shao2021adversarial,mahmood2021robustness,paul2021vision,naseer2021intriguing,aldahdooh2021reveal,bhojanapalli2021understanding} concurrently compare the robustness of transformers to CNNs and independently derive conclusions resembling each other. Their main conclusions, ignoring their nuance difference, can be summarized as \textit{vision transformers are more robust than CNNs}. Not surprisingly, our work also comes to the same main take-way message but differs in multiple aspects, such as co-analysis of MLP-Mixer, perturbation-minimization (C\&W and DeepFool) results in white-box setting, evaluation under universal attack. Our works also leads to some additional insight, such that MLP-Mixer shows an increased vulnerability to universal attacks. In another parallel line,~\cite{mao2021rethinking,naseer2021improving} have investigated how to improve the adversarial robustness of vision transformers. A future version will further discuss the detailed differences among the above concurrent works.

\section{Research goal and Experimental setup}
{\bf Research goal and scope.} CNNs have achieved dominant success in numerous vision applications in the last few years, however, they are also vulnerable to adversarial attacks. Such vulnerability causes serious concern in security-sensitive applications, such as autonomous driving. This concern has motivated an extensive study on model robustness against various attack methods. With the recent popularity of ViT and Mixer as alternatives to CNNs, it is vital for the community to understand their adversarial robustness and to benchmark them against the widely used CNNs.
To this end, this work \textbf{\textit{empirically}} investigates the adversarial robustness of the three architectures. In other words, this work has no intention to understand the reason behind why a certain architecture is more or less robust. Note that there is still no consensus~\cite{akhtar2018threat} on the explanation of CNN being sensitive to adversarial examples despite a large body of works in this field. As an early attempt to investigate the adversarial robustness of ViT and Mixer, our work focuses on the empirical evaluation and it is out of the scope of this work to theoretically understand why they might be vulnerable. Nonetheless, our work attempts to provide better understandings of the robustness gap between models from a shift-invariance perspective and a frequency perspective. Admittedly, our attempt for explanation is limited and future work is needed for better understanding. 

{\bf Models and dataset.}
In our experiments, we mainly compare the ViT~\cite{dosovitskiy2021an} models, MLP-Mixer~\cite{tolstikhin2021mlpmixer} and CNN architectures~\cite{he2016deep}. Note that they all all adopt shortcut~\cite{he2016deep} in their architecture design. For the ViT models, we consider ViT-B/16 and ViT-L/16, where B and L stand for ``base" and ``large", respectively, while $16$ indicates the patch size. The considered ViT models were pre-trained on ImageNet-21K and fine-tuned on ImageNet-1K~\cite{deng2009imagenet}.
We also evaluate ViT models that are directly trained on ImageNet-1K from~\cite{steiner2021train} (indicated by ``ImageNet-1K"). 
Corresponding to the ViT models, we also investigated Mixer-B/16 and Mixer-L/16~\cite{tolstikhin2021mlpmixer} which are trained on ImageNet-1K. We further consider CNN architectures, ResNet-18 and ResNet-50~\cite{he2016deep} trained on ImageNet-1K as well as the semi-weakly supervised (SWSL) variant~\cite{yalniz2019billion}, which is
pre-trained on IG-1B-Targeted~\cite{mahajan2018exploring} with associated hashtags from 1,000 ImageNet-1K classes followed by fine-tuning on ImageNet-1K.
To evaluate adversarial attacks, if not otherwise mentioned we evaluate different adversarial attacks in the untargeted setting on an ImageNet-compatible dataset. This dataset was originally introduced in the NeurIPS 2017 adversarial challenge\footnote{\url{https://github.com/rwightman/pytorch-nips2017-adversarial}}. 
We compare different architectures for the corresponding most widely used models, such as ResNet-18/50 for CNN, and ViT and Mixer for B/16 and-L/16. However, we also note that many other factors other than the architecture itself might also have a role in influencing the robustness. Given the publicly available models, it is nearly impossible to rule out all other factors.

\section{Experiment Results}
\subsection{Robustness Against White-Box Attacks}
We first investigate the robustness under white-box attacks. Particularly, we deploy PGD~\cite{madry2017towards} and FGSM~\cite{goodfellow2014explaining}. For both attacks we consider $\epsilon=\{d/255\mid d\in\{0.1$, $0.3$, $0.5$, $1$, $3\}\}$ for images in range $[0,1]$. For the PGD attack, we set the number of iterations to $20$ and keep the other parameters as the default settings of Foolbox~\cite{rauber2017foolboxnative}. For these two attacks, we report the attack success rate (ASR), \ie the percentage of samples which were classified differently from the ground-truth class. Additionally, we evaluate the models on the $\ell_2$-variants of the C\&W attack~\cite{carlini2017towards} and DeepFool~\cite{moosavi2016deepfool}. These two attacks have the objective to minimize the perturbation magnitude given the ASR of 100\%. Hence, we report the $\ell_2$-norm 
of the adversarial perturbation and the results are available in Table~\ref{tab:white_box_robustness}. Overall a trend can be observed that compared with CNN architectures, the ViT and Mixer models have a lower attack success rate, suggesting they are more robust than CNN architectures. The increased robustness of ViT and Mixer models is further supported by a higher $\ell_2$-norm for the C\&W and DeepFool attacks. However, when the perturbation magnitude is very small, the opposite phenomenon can be observed. For example for Mixer-L/16 for the PGD or FGSM with $\epsilon=0.1$ the Mixer and ViT models can exhibit decreased robustness compared to the CNN models. 

\begin{figure}[!htbp]
\includegraphics[width=0.23\linewidth]{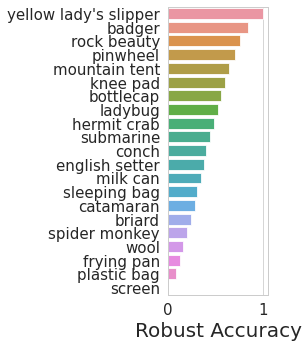}
\includegraphics[width=0.37\linewidth]{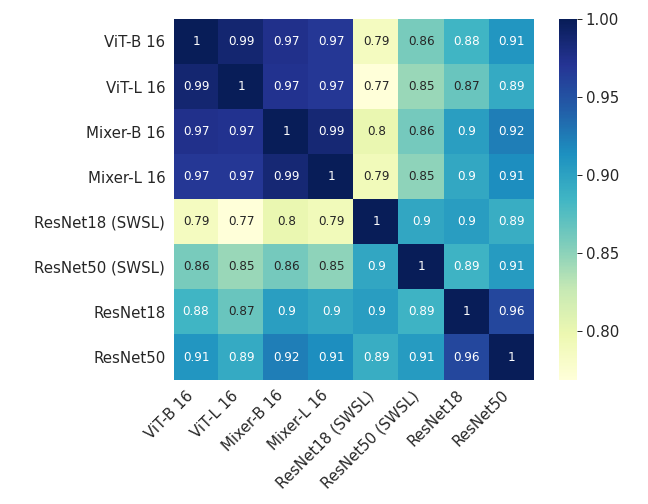}
\includegraphics[width=0.37\linewidth]{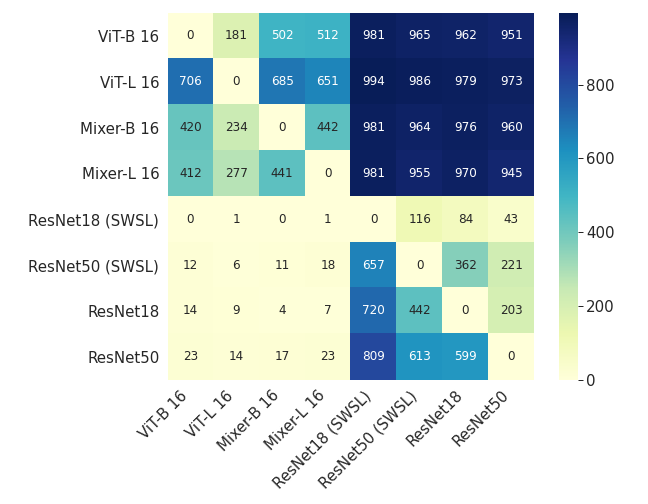}
\caption{\textbf{(Left):} Most to least robust classes in increments of $50$ for ViT-L/16; \textbf{(Center):} Cosine similarity of class-wise robust accuracies between different models; \textbf{(Right):} Count of all class-wise accuracies of one model (row) which are more robust than those of another (column). The robustness of all models is evaluated under the $\ell_\infty$-PGD attack with $\epsilon = 0.3$ on the ImageNet validation dataset.}
\label{fig:class-wise-robustness}
\end{figure}

\begin{table*}[!htbp]
\begin{center}
\scalebox{0.55}{
\begin{tabular}{|l|c|c|c|c|c|c|c|c|c|c|c|c||c|c|}
\hline
 & \multicolumn{2}{c|}{Clean} & \multicolumn{5}{c|}{PGD ($\ell_\infty$) $\downarrow$} & \multicolumn{5}{c|}{FGSM ($\ell_\infty$) $\downarrow$}  & \multicolumn{1}{|c}{C\&W ($\ell_2$)$\uparrow$} & DeepFool ($\ell_2$)$\uparrow$ \\\hline
Model
& ImageNet & NeurIPS & 0.1 & 0.3&0.5 & 1 & 3 & 0.1 &0.3& 0.5 & 1 & 3  &&\\ \hline\hline
ViT-B/16 
& 81.4 & 90.7 & \textbf{22.6} & 63.6 &86.5  & 97.5 & 99.9 & \textbf{19.1} &38.7&52.8  & 66.3 & 79.7  & \textbf{0.468} & 0.425 \\
ViT-L/16 
& 82.9 & 89.3 & 22.8 &\textbf{60.1}& 80.9 & 95.8 & 100 & 19.5 &\textbf{35.9}& \textbf{44.9}     & \textbf{57.9} & \textbf{67.3} & 0.459 & \textbf{0.548} \\
ViT-B/16 (ImageNet-1K) & 76.7 & 85.6 & 29.2 & 68.3 & 87.1 & 96.8 & 99.5 & 26.7 & 52.2 & 67.1 & 83.4 & 91.0 &0.408	&0.308 \\ 
ViT-L/16 (ImageNet-1K) & 72.8 & 79.2 & 44.9 & 77.0 & 90.6 & 97.1 & \textbf{99.2} & 36.8 & 51.4 & 59.8 & 70.3 & 78.8 &0.335 &0.261\\ 
Mixer-B/16 
& 76.5 & 86.2 & 29.5 &63.4&82.0  & 96.2 & 100 & 27.7 &49.3& 59.5     & 69.3 & 78.0 & 0.375 & 0.339 \\
Mixer-L/16 
& 71.8 & 80.0 & 41.1 &67.3&\textbf{80.4}  & \textbf{92.1} & 99.4& 36.7 &51.8& 56.9    & 61.6 & 67.4 & 0.297 & 0.377 \\
ResNet-18 (SWSL) 
& 73.3 & 90.4 & 47.9 &93.7&98.7  & 99.5 & 99.6 & 38.0 &76.3& 89.9     & 96.2 & 97.6  & 0.295 & 0.132 \\
ResNet-50 (SWSL) 
& 81.2 & 96.3 & 39.4 &90.2&97.0  & 98.4  &99.4& 26.3 &60.9& 73.0    & 83.8 & 87.5  & 0.380 & 0.149 \\
ResNet-18
& 69.8 & 83.7 & 46.1 &90.0&97.8 & 99.9 & 100 & 42.0&75.2 &88.5  & 95.7 & 98.2 & 0.302 & 0.237 \\
ResNet-50
& 76.1 & 93.0 & 35.8 &86.3 &97.9  & 99.5 & 100 & 27.5 &63.1&77.6  & 89.4 & 93.9 & 0.371 & 0.287 \\ 
\hline
\end{tabular}
}
\end{center}
\vspace{-0.5\baselineskip}
\caption{\textbf{White-box attacks on benchmark models with different epsilons.} We report the clean accuracy on the ImageNet and NeurIPS dataset. The attack success rate (\%) of PGD and FGSM under $\ell_\infty$ distortion, and the $\ell_2$-norm of C\&W and DeepFool, respectively are reported for the NeurIPS dataset. All models are trained with an image size of $224$. A model with a \textbf{lower ASR$\downarrow$} or \textbf{higher $\ell_2$-norm$\uparrow$} is considered to be more robust.} 
\label{tab:white_box_robustness}
\end{table*}
\textbf{Class-wise Robustness.} To provide a more detailed robustness evaluation, we perform a class-wise robustness study. We perform the class-wise robustness study on the ImageNet validation dataset, where for each class $50$ validation images are present and attack the models with the $\ell_\infty$-PGD attack ($\epsilon=0.3$). 
Figure~\ref{fig:class-wise-robustness} (left) shows every 50th class from the most robust classes to the least robust classes for ViT-L/16, where the class ``screen" exhibits no robustness, while the class ``yellow lady's slipper" exhibits robustness of $100\%$. This indicates an imbalance in the class-wise robustness between different classes. We further calculate the similarity of the class-wise accuracies between different models. We treat the different class-wise accuracies as a vector and calculate the cosine similarity between the class-wise accuracies of different models. From Figure~\ref{fig:class-wise-robustness} (center), it can be observed that the ViT and Mixer models exhibit relatively high similarity values, but also that ResNet18 and ResNet50 are similar in their class-wise robust accuracies. Finally, we examine the relative class-wise robustness of the models, by calculating how many classes of one model are more robust than those of another model. These results are presented in Figure~\ref{fig:class-wise-robustness} (right). Here it can be observed that the ViT and Mixer models exhibit higher class-wise accuracies than the CNNs, with consistently more than $945$ classes being more robust than the examined CNNs. 

\subsection{Robustness Against Black-Box Attacks}
We evaluate and compare the robustness of different model architectures for the black-box attacks in two setups: query-based black-box attack and transfer-based black-box attack.
\begin{table}[!htbp]
\centering
\resizebox{0.7\linewidth}{!}{%
\begin{tabular}{|l|c|c|c|c|c|c|c|c|c|c|}
\hline
 &
  ViT-B &
  ViT-L &
  Mixer-B &
  Mixer-L &
  \begin{tabular}[c]{@{}c@{}}RN18 (SWSL)\end{tabular} &
  \begin{tabular}[c]{@{}c@{}}RN50 (SWSL)\end{tabular} &
  RN18 &
  RN50 
  \\ \hline\hline
\begin{tabular}[c]{@{}c@{}}Boundary\\ ($\ell_2$)\end{tabular} &
  3.980 &7.408 &1.968 & 1.951&  1.403 & 1.846 
  &  1.468 &  1.740\\ \hline
\end{tabular}%
}
\caption{\textbf{Boundary attack on benchmark models}. We test 100 randomly selected images from NeurIPS dataset against decision-based attack, and the $\ell_2$-norm of adversarial perturbation is presented.}
\label{tab:tab_query}
\end{table}

\begin{table}[!htpb]
\centering
\resizebox{1\linewidth}{!}{%
\begin{tabular}{|l|cc|cc|cc|cc|cc|cc|cc|cc|cc|cc|}
\hline
 & \multicolumn{2}{c|}{ViT-B/16} & \multicolumn{2}{c|}{ViT-L/16} & \multicolumn{2}{c|}{Mixer-B/16} & \multicolumn{2}{c|}{Mixer-L/16} & \multicolumn{2}{c|}{RN18 (SWSL)} & \multicolumn{2}{c|}{RN50 (SWSL)} & 
 \multicolumn{2}{c|}{RN18} & \multicolumn{2}{c|}{RN50} \\\hline
 Methods & ASR & \#Q & ASR & \#Q & ASR & \#Q & ASR & \#Q & ASR & \#Q 
 & ASR & \#Q & ASR & \#Q & ASR & \#Q \\\hline\hline
{\bf Bandits$_{TD}$ ($\ell_2$)} &91\%  & 1307 &81\%  &\textbf{2558}  &41\% &1748 &87\%  &1697  &97\%  &564  &85\%  &893 
&98\%  &861  &94\%  &743 \\
\hline
\end{tabular}%
}
\caption{{\bf Results of Bandits$_{TD}$ (with time and data-dependent priors) attack on benchmark models.} ASR is the attack success rate and \#Q denotes the average number of queries of the successful attacks. We use 100 randomly selected images from NeurIPS dataset with a query limit of 10,000.}
\label{tab:tab_query2}
\end{table}

{\bf Query-based black-box attacks.} 
Query-based black-box attacks work by evaluating a sequence of perturbed images through the model. 
We adopt a popular decision-based attack, the Boundary Attack~\cite{brendel2017decision}, which only requires a model's final decision (\ie, class label) and aims to minimize the perturbation while remaining adversarial. 
As with the white-box attack, a trend can be observed in the black-box attack that the ViT and Mixer models are more robust, indicated by the relatively higher $\ell_2$-norm of the adversarial perturbation (see Table~\ref{tab:tab_query}). We further test and compare the models using a more recently proposed
approach~\cite{ilyas2018prior} that reduces the query cost.
To achieve a higher success rate with fewer queries, 
Bandits$_{TD}$ integrates the prior information about the gradient (\ie time and data) using bandit algorithm to reduce the query number. Following the setting in~\cite{ilyas2018prior}, the maximum $\ell_2$-norm of perturbations is set to 5 and the others are set to default as well. As shown in Table~\ref{tab:tab_query2}, overall we observe that ViT and Mixer require a larger average number of queries with a lower average ASR, suggesting ViT and Mixer are more robust than their CNN counterparts.
\newline
{\bf Transfer-based black-box attacks.} Transfer-based black-box attacks exploit the transferable property of adversarial examples, \ie, the adversarial examples generated on a source model transfer to another unseen target model. For the source model, we deploy the I-FGSM~\cite{kurakin2016adversarial} attack with 7 steps and evaluate the transferability on the target model.
From the result in Table~\ref{tab:blackbox}, we have two major observations. First, adversarial examples from the same family (or similar structure) exhibit higher transferability, suggesting models from the same family learn similar features. Second, when a different model architecture is used as the source model, there is also a trend that CNNs are relatively more vulnerable (\ie, transfer poorly toward foreign architectures). For example, the transferability from CNN to ViT is often lower than 20\%, while the opposite scenario is higher.

\begin{table*}[!htpb]
\begin{center}
\scalebox{0.63}{
\begin{tabular}{|l|c|c|c|c|c|c|c|c|c|}
\hline 
& \multicolumn{1}{c|}{} 
 & \multicolumn{8}{c|}{\textbf{Target model}}  \\ \hline
 \multicolumn{1}{|c|}{\textbf{Source model}} &
\multicolumn{1}{c|}{\textbf{Variant}}  &
  \textbf{ViT-B/16} &
  \textbf{ViT-L/16} &
  \textbf{Mixer-B/16} &
  \textbf{Mixer-L/16} &
  \textbf{ResNet-18} &
  \textbf{ResNet-50} &
  \textbf{ResNet-18} &
  \textbf{ResNet-50} \\ 
  &&&&&&(SWSL)&(SWSL)&&
  \\\hline\hline
ViT-B/16                            & I-FGSM  & 100  & 84.7 & 48.8 & 50.5 & 32.0 & 20.5 & 40.9 & 31.7  \\
ViT-L/16                            & I-FGSM  & 90.9 & 99.9 & 45.7 & 48.0 & 30.4 & 22.2 & 40.8 & 30.9 \\
Mixer-B/16                          & I-FGSM  & 33.9 & 25.3 & 100  & 89.1 & 30.6 & 20.5  & 40.8 & 32.0 \\
Mixer-L/16                          & I-FGSM  & 27.7 & 20.1 & 80.3 & 99.7 & 27.7 & 17.0 & 38.2 & 28.4 \\
ResNet-18 (SWSL)                    & I-FGSM  & 16.2 & 13.6 & 24.8 & 29.5 & 99.6 & 57.1  & 73.5 & 63.4 \\
ResNet-50 (SWSL)                    & I-FGSM  & 15.3 & 13.5 & 23.6 & 29.9 & 56.5 & 99.5 & 49.4 & 51.0 \\
ResNet-18                          & I-FGSM  & 18.2	&14.7	&28.9	&35.6	&84.6	& 49.9 & 100	&81.6 \\
ResNet-50                          & I-FGSM  & 17.7	&13.6	&28.4	&34.5	&73.9	&63.9 & 80.6 & 100  \\ 
\hline
\end{tabular}%
}
\end{center}
\vspace{-0.5\baselineskip}
\caption{\textbf{Transfer-based black-box attacks on benchmark models.} We report the attack success rate (\%) and a model with a lower ASR is considered to be more robust. All models are trained with an image size of $224$, and attacked with a maximum $\ell_\infty$ perturbation of $\epsilon=16$.}
\label{tab:blackbox}
\end{table*}

\subsection{Toy Example}
\begin{wrapfigure}[6]{r}{0.4\linewidth}
\vspace{-5mm}
	\centering
	\includegraphics[width=0.53\linewidth]{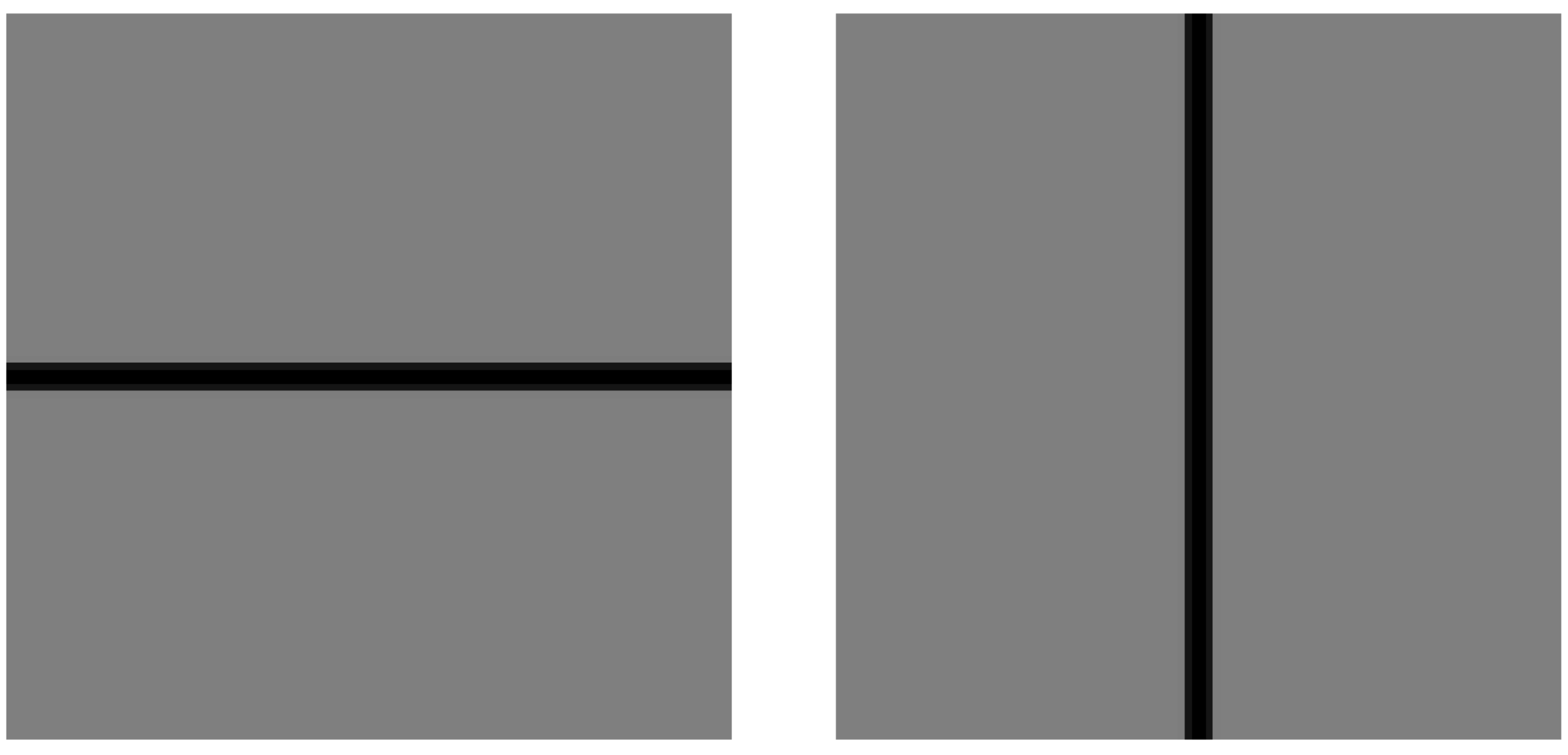}
	\vspace{-0.5\baselineskip}
	\caption{Images for our binary classification toy example.}
	\label{fig:toy_example_images}
\end{wrapfigure}
ViT and Mixer are more robust to adversarial attacks than conventional CNNs. In other words, CNN tends to be the least robust in most setups. To facilitate the understanding of the mechanisms, we design a toy example of binary classification where each class is represented by a single image with a size of 224. The two images consist of a single black stripe on a grey background, differing in the stripe orientation, namely a vertical and a horizontal stripe. The two images used for training are shown in Figure~\ref{fig:toy_example_images}. 

\begin{wraptable}[5]{r}{0.43\linewidth}
\vspace{-4mm}
\centering
\scalebox{0.6}{
\begin{tabular}{|l|c|c|c|c|}
\hline
    & C\&W ($\ell_2$)       & DDN ($\ell_2$)                 &$\#params$\\ \hline\hline
CNN & 12.55                  & 17.36      &$4.59M$\\
FC  & 25.06                 & 25.39       &$4.82M$\\
ViT & 27.82                 & 59.99       &$4.88M$\\ \hline
\end{tabular}
}
\vspace{0.5\baselineskip}
\caption{$\ell_2$-norm of the adversarial perturbation on our toy example.}
\label{tab:toy_example_result}
\end{wraptable}
\noindent We then train a Fully Connected network (FC), a Convolution Neural Network (CNN), and a Vision Transformer (ViT) on the images. Note that we designed the networks to be of relatively small capacity ($<5M$), due to the simplicity of the task and to constrain that the networks have around the same number of parameters. We evaluate the adversarial robustness of these models with the commonly used $\ell_2$ attacks C\&W~\cite{carlini2017towards} and DDN~\cite{rony2019decoupling}. We report the $\ell_2$-norm of the adversarial perturbation in Table~\ref{tab:toy_example_result}. It can be observed that the CNN is also less robust than the FC and the ViT in this toy example setup. 
\newline
\begin{figure}[!htbp]
\centering
\includegraphics[width=0.77\linewidth]{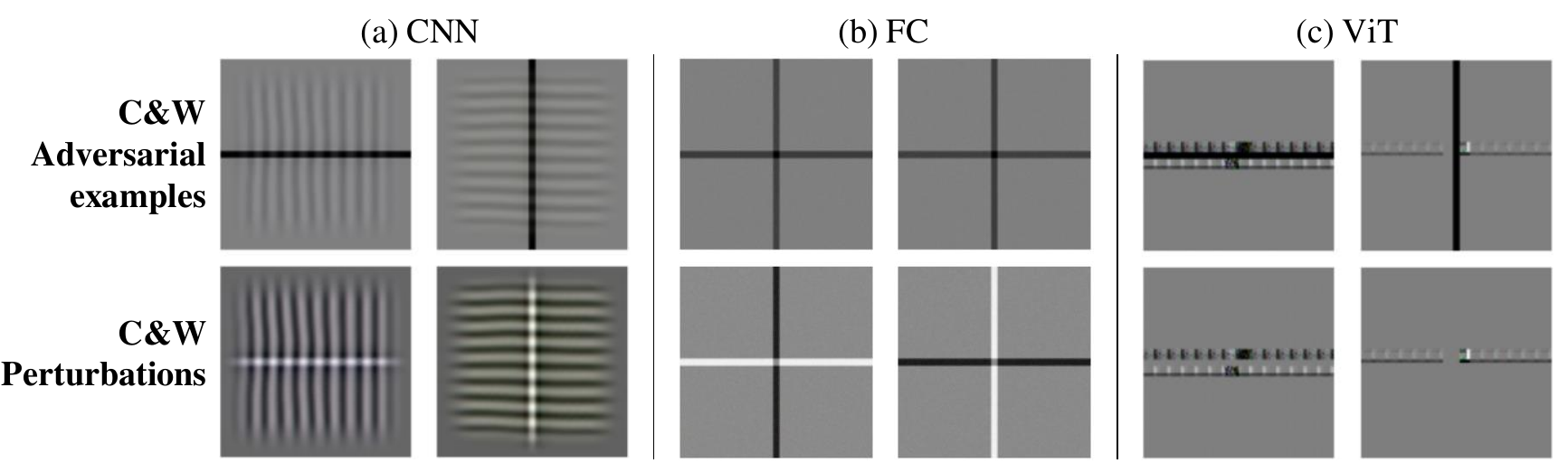}
\includegraphics[width=0.77\columnwidth]{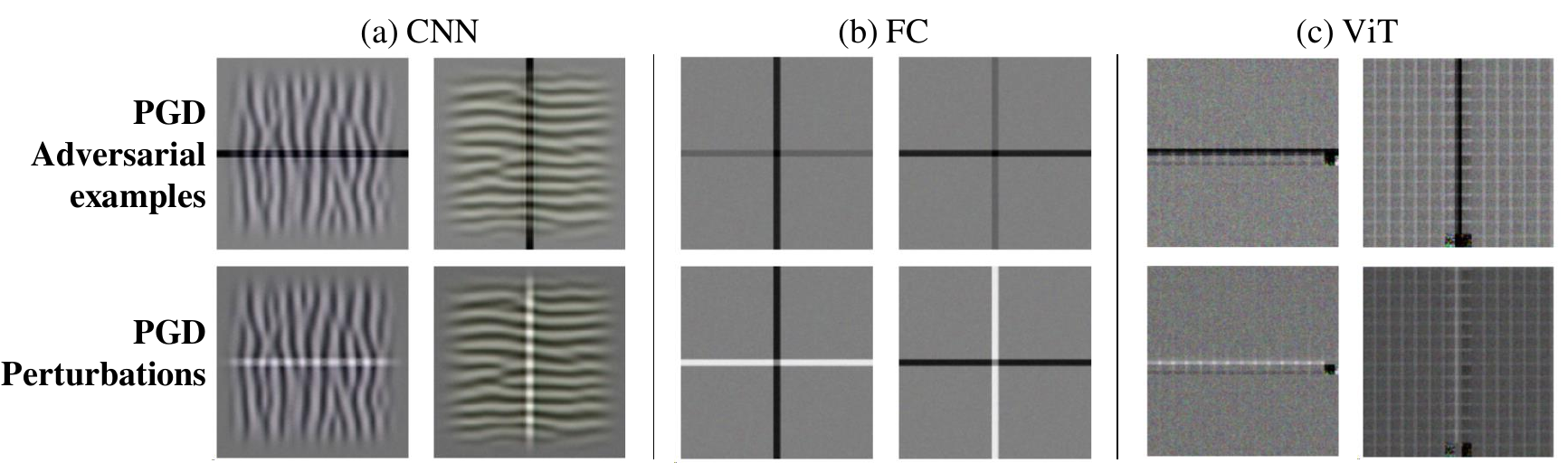}
\vspace{-0.5\baselineskip}
\caption{Adversarial examples and perturbations generated with C\&W (top) and PGD (bottom) using different architectures trained on our toy example.}
\label{fig:toy_example_adv}
\end{figure}

\textbf{Explanation from the perspective of shift-invariance.} Recently,~\cite{ge2021shift} has shown that the shift-invariance property of CNNs can be one cause for its vulnerability to adversarial attacks. Their conclusion is derived with the quantitative analysis and theoretical proof, while we focus on providing an intuitive qualitative analysis with the observation that shift-invariance property causes adversarial examples with a repetitive pattern. The qualitative results of adversarial perturbations generated by the attacks are shown in Figure~\ref{fig:toy_example_adv}. For the ViT, one phenomenon can be observed that the adversarial perturbation consists of square patches. This is likely due to the division of the input image into patches in the ViT architecture. Without this split process on the image, we observe clear stripes but with different patterns for CNN and FC. While the CNN model generates perturbations with repeated stripes, the FC model generates perturbations with only a  single stripe centered in the image. It should be noted that perturbations are generated toward the adversary, \ie, in the direction of the opposite class' stripe. 
The observation that the CNN model yields stripes all over the image can be attributed to the shift-invariant property of the CNN model. From the perspective of shift-invariance, the CNN model recognizes features, \ie horizontal or vertical stripe in this setup, regardless of the position of the features on the image. Thus, it is somewhat expected that the perturbation has stripes in a different direction all over the image. For the FC model without the shift-invariant property, it only recognizes the stripes in the center; thus, the resulting perturbation mainly has the stripe in the center. 
Since our toy example only consists of two samples, the ASR has only limited informative value about the model's robustness. However, the qualitative results can still be observed. For the $\ell_2$-PGD attack, we choose a sufficiently high $\epsilon$ of $40$, such that both samples are misclassified. The qualitative results of the PGD attack in Figure~\ref{fig:toy_example_adv} (bottom) resemble those of the C\&W attack. These qualitative results provide an interesting insight into the possible link between shift-invariant property and CNN vulnerability. Admittedly, this link is vague and future work is needed to establish a more concrete link between them.

\subsection{Frequency Analysis}
\begin{figure}[!htbp]
	\centering
	\includegraphics[width=0.85\columnwidth]{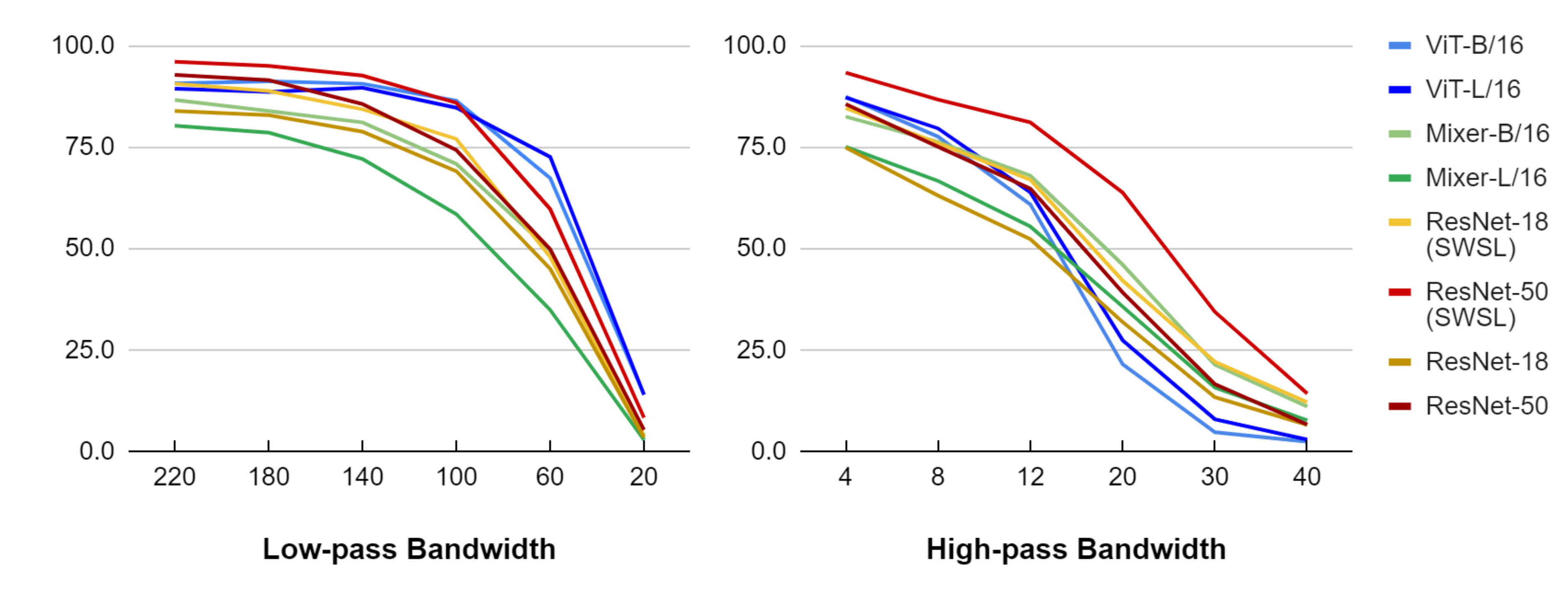}
    \vspace{-0.5\baselineskip}
	\caption{Top-1 accuracy across a range of frequency bandwidths from low/high-pass filtering (The right direction of x-axis is more oriented to the low and high frequency, respectively). \textbf{Left:} Low-pass filtering. \textbf{Right:} High-pass filtering.}
    \label{fig:freq}
\end{figure}

We further attempt to explain the lower robustness of CNN from a frequency perspective~\cite{yin2019fourier,zhang2021universal}. Following~\cite{yin2019fourier,zhang2021universal}, we deploy a low-pass filter to filter out high-frequencies and a high-pass filter to filter out low-frequencies from the input images before feeding them to the model. We then evaluate the Top-1 accuracy of images from the NeurIPS dataset by applying low-pass or high-pass filtering, and the results are shown in Figure~\ref{fig:freq}.
For the low-pass filtering, a sharper decline of the CNN architectures can be observed than for the ViT and Mixer, indicating that the CNN architectures are more reliant on the high-frequency features compared to other models.
For example, ResNet-50 (SWSL) has higher accuracy than ViT-L/16 when filtered with large bandwidths; however, with smaller bandwidths, the accuracy of ViT-L/16 becomes higher than the ResNet. Additionally, the Mixers have a sharper decline than the ViTs at low-pass bandwidths from 60 to 180. For the high-pass filtering, the ViT models show the steepest decline among the models, indicating that the ViT models rely more on low-frequency features while CNNs relatively are more biased towards low-frequency features. Note that non-robust features tend to have high-frequency properties~\cite{yin2019fourier,zhang2021universal,ilyas2019adversarial} and explain the decreased model robustness. This demonstrates why ViT models are more robust than CNN architectures from the frequency perspective. Comparing results from both low-pass and high-pass filtering, we observe that, regardless of their absolute value of accuracy, Mixers exhibit a similar trend to CNNs rather than ViTs.

\section{Additional Investigations}
\subsection{Robustness against common corruptions} 
Additional to our investigation of the adversarial vulnerability of ViT, Mixer, and CNN models, we examine the robustness of these models to common, \ie naturally occurring, corruptions~\cite{hendrycks2019benchmarking}. ImageNet-C was proposed by~\cite{hendrycks2019benchmarking} benchmarking neural network robustness against these common corruptions. In essence, ImageNet-C is a perturbed version of the original ImageNet validation dataset that has 1000 classes and 50 images per class. Specifically, ImageNet-C has 15 test corruptions, each corruption type having 5 different severities, and four hold-out corruptions. 

\begin{figure}[!htbp]
	\centering
	\includegraphics[width=0.82\columnwidth]{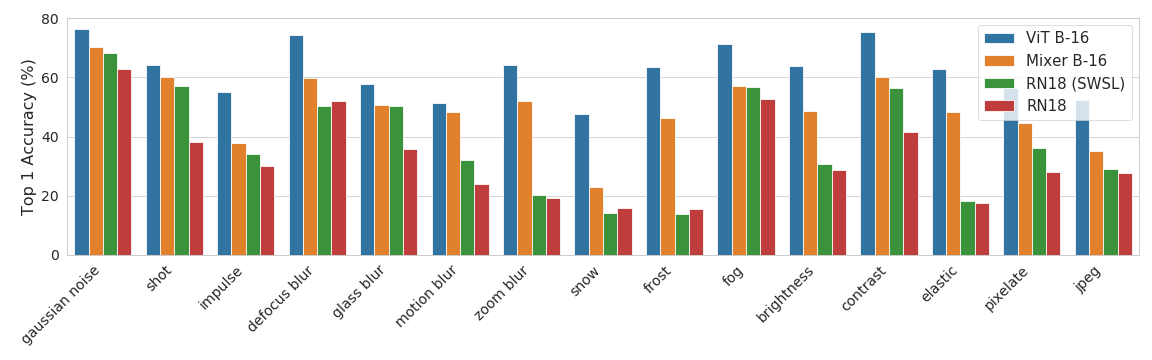}
    \vspace{-0.8\baselineskip}
	\caption{Detailed corruption performance comparison of the different models.} 
    \label{fig:corruption}
\end{figure}

\begin{wraptable}[9]{r}{0.38\linewidth}
\vspace{-8mm}
\centering
\scalebox{0.58}{
\begin{tabular}{|l|c|c|c|}
\hline
            & ImageNet & ImageNet-C & mCE   \\ \hline\hline
ViT-B/16    & 81.43    & 58.85      & 51.98 \\
ViT-L/16    & 82.89    & 64.11      & 45.46 \\
Mixer-B/16  & 76.47    & 47.00      & 67.35 \\
Mixer-L/16  & 71.77    & 40.47      & 75.84 \\
RN18 (SWSL) & 73.29    & 37.84      & 78.30 \\
RN50 (SWSL) & 81.18    & 52.03      & 60.49 \\
RN18        & 69.76    & 32.92      & 84.67 \\
RN50        & 76.13    & 39.17      & 76.70 \\
\hline
\end{tabular}
}
\vspace{0.5\baselineskip}
\caption{\textbf{Evaluation of benchmark models on the ImageNet-C dataset.} Accuracy the higher the better, mCE the lower the better.}
\label{tab:corruption_robustness}
\end{wraptable}
\noindent Following~\cite{hendrycks2019augmix}, we evaluate on 15 test corruptions and the results are shown in Table~\ref{tab:corruption_robustness}. First, there is a clear trend that models that have higher accuracy on the original (clean) ImageNet tend to also have higher accuracy on ImageNet-C, which is somewhat expected. Second, with comparable accuracy on the original (clean) ImageNet, ViT, and Mixer architectures tend to have higher robustness against corruptions. For example, ViT-B/16 has similar accuracy as RN50(SWSL), \ie 81.43\% \vs 81.18\%, but the robustness of ViT-B/16 is notably higher than that of RN50(SWSL), \ie 58.85\% \vs 52.03\%. A similar phenomenon can be observed for Mixer-B/16 and RN50. More detailed corruption-wise results are shown in Figure~\ref{fig:corruption}. We find that under some corruptions, such as zoom blur and snow, the superiority of ViT is more significant than other corruptions, such as Gaussian noise.

\subsection{Robustness against Universal Adversarial Perturbations}
\begin{figure}[!htbp]
	\centering
	\includegraphics[width=0.6\linewidth]{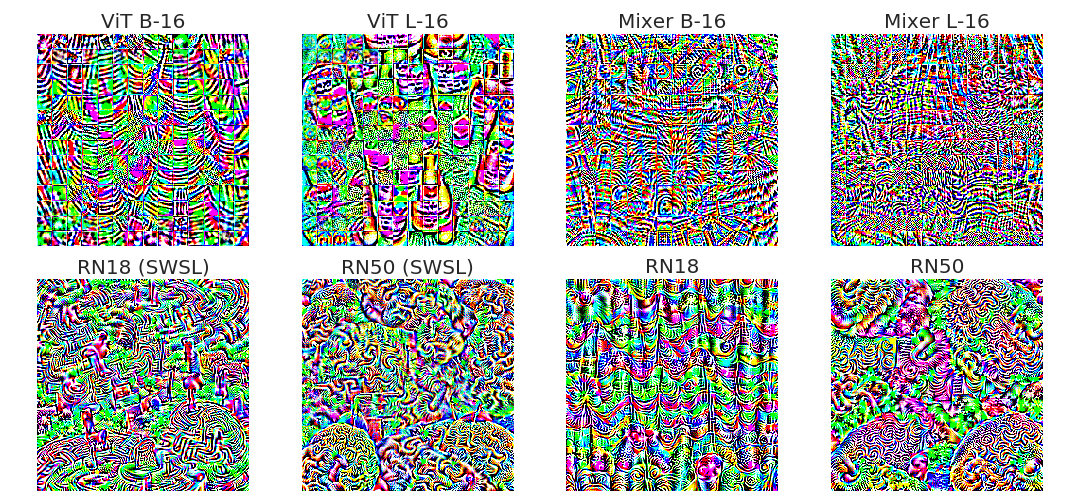}
	\vspace{-0.5\baselineskip}
	\caption{Visualization of the amplified UAPs generated on different model architectures. }
	\label{fig:uaps}
\end{figure}

We further compare the robustness of the different model architectures against Universal Adversarial Perturbations (UAPs). UAPs have been proposed in~\cite{moosavi2017universal}, however, their algorithm has been identified to be relatively less effective yet slow. In this work, we adopt the state-of-the-art UAP algorithm in~\cite{zhang2020understanding}. Following~\cite{moosavi2017universal,zhang2020understanding}, we set the image size to $224$ and constrain the UAP with an $\ell_\infty$-norm of $\epsilon=10/255$. The white-box and the corresponding transferability results are shown in Table~\ref{tab:uap}. Several observations can be made. First, MLP-Mixer models are very vulnerable to UAPs in both white-box and black-box scenarios. Especially in the black-box scenario, the attack success rate is always higher than 95\% regardless of the surrogate model architecture. On the contrary, ViT models and CNN models are more robust against UAPs, particularly in the black-box scenarios. Second, comparing ViT and CNN models, there is no obvious robustness gap with an ASR of around 90\% for all models. However, in the more challenging black-box scenarios, the ViT models are significantly more robust than their CNN counterparts. This trend aligns well with our previous finding that ViT models are more robust than CNN models. We visualize amplified versions of the resulting UAPs in Figure~\ref{fig:uaps}. It is noticeable that for the ViT and Mixer architectures, a tile pattern is observable, which is caused by the operation of dividing the images into patches as tokens. Another interesting observation is that the UAPs of Mixer tend to be less locally smooth than those generated on ViT and CNN. ViT-L/16 with the highest robustness against UAP seems to also have the most locally smooth patterns. Overall, it is interesting that Mixer is extremely vulnerable to UAPs and the finding from a qualitative result that they have locally non-smooth patterns. However, we have no clear explanation for the observed phenomenon since the Mixer is still a very recent architecture. An important message to the community is that the adversarial threat to Mixer cannot be ignored since it is vulnerable in the \textit{practical} transferable universal attack scenario.
\begin{table*}[!htpb]
\centering
\scalebox{0.58}{
\begin{tabular}{|l|c|c|c|c|c|c|c|c|c|c|c|c|c|c|}
\hline

 & \multicolumn{8}{c|}{\textbf{Target model}}  \\ \hline
 \textbf{Source model}& ViT-B/16 & ViT-L/16 & Mixer-B/16 & Mixer-L/16 & RN18 (SWSL) & RN50 (SWSL) & RN18 & RN50 \\ \hline\hline
ViT-B/16    &\textbf{93.9} & 32.7 & 95.7 & 96.0   & 43.3 & 29.2 & 46.4 & 37.2 \\
ViT-L/16    &42.1 & \textbf{84.8} & 95.6 & 95.6 & 39.0   & 21.2 & 41.6 & 31.7 \\
Mixer-B/16  &16.6 & 14.9 & \textbf{98.6} & 96.7 & 35.4 & 19.6 & 43.3 & 30.6 \\
Mixer-L/16  &15.2 & 13.7 & 96.3 & \textbf{98.9} & 30.3 & 15.5 & 38.7 & 26.0   \\
RN18 (SWSL) &12.6 & 12.1 & 95.4 & 95.6 & \textbf{91.2} & 49.2 & 56.6 & 56.5 \\
RN50 (SWSL) & 11.8 & 11.9 & 95.3 & 95.6 & 50.4 & \textbf{87.1} & 46.8 & 46.6 \\
RN18        &13.5 & 11.6 & 95.7 & 95.7 & 66.2 & 46.1 & \textbf{93.7} & 63.3 \\
RN50        &12.2 & 12.0 & 95.4 & 95.6 & 63.3 & 57.0 & 63.4 & \textbf{91.9} \\
\hline
\end{tabular}
}
\vspace{0.5\baselineskip}
\caption{{\bf Universal adversarial attacks on benchmark models.} We report the attack success rate (\%) and a model with a lower ASR is considered to be more robust. All models are trained with an image size of $224$, and evaluated on the NeurIPS dataset. The \textbf{bold} font indicates white-box attacks.}
\label{tab:uap}
\end{table*}

\section{Conclusion}
Our work performs an empirical study on the adversarial robustness comparison of ViT and MLP-Mixer to the widely used CNN on image classification. Our results show that ViT is significantly more robust than CNN in a wide range of white-box attacks. A similar trend is also observed in the query-based and transfer-based black-box attacks. Our toy task of classifying two simple images with a vertical or horizontal black stripe in the image center provides interesting insight on the possible link between shift-invariant property and CNN vulnerability and future work is necessary for further investigating this link. Our analysis from the feature perspective further suggests that ViTs are more reliant on low-frequency (robust) features while CNNs are more sensitive to high-frequency features. We also investigate the robustness of the newly proposed MLP-Mixer and find that its robustness generally locates in the middle of ViT and CNN. We have also performed additional investigations on robustness against common corruptions and UAPs. One very intriguing finding is that Mixer is extremely vulnerable to UAPs, even in transfer-based black-box attacks. Future work is needed for a better understanding of the reported empirical results.


\bibliography{bib_mixed}
\end{document}